\documentclass[conference]{IEEEtran}
\ifCLASSINFOpdf
  \usepackage[pdftex]{graphicx}
  \graphicspath{{../figures/}}
  \DeclareGraphicsExtensions{.pdf,.jpeg,.png,.eps}
\else
   \usepackage[dvips]{graphicx}
  \graphicspath{{../figures/}}
  \DeclareGraphicsExtensions{.eps}
\fi
\usepackage{amsmath}
\usepackage{amssymb}
\ifCLASSOPTIONcompsoc
  \usepackage[caption=false,font=normalsize,labelfont=sf,textfont=sf]{subfig}
\else
  \usepackage[caption=false,font=footnotesize]{subfig}
\fi
\usepackage{url}
\usepackage{xcolor}

\begin{document}
%
% paper title
% Titles are generally capitalized except for words such as a, an, and, as,
% at, but, by, for, in, nor, of, on, or, the, to and up, which are usually
% not capitalized unless they are the first or last word of the title.
% Linebreaks \\ can be used within to get better formatting as desired.
% Do not put math or special symbols in the title.
\title{Image operator learning coupled with CNN classification and its
  application to staff line removal}

% author names and affiliations
% use a multiple column layout for up to three different
% affiliations
%\author{\IEEEauthorblockN{Author names omitted for double-blind review}}
 \author{\IEEEauthorblockN{Frank D. Julca-Aguilar and Nina S. T. Hirata}
   \IEEEauthorblockA{Department of Computer Science, Institute of
     Mathematics and Statistics\\
     University of S\~ao Paulo (USP)\\
     S\~ao Paulo, Brazil}
}
% \\Email: \{faguilar, nina\}@ime.usp.br}}
% \and
% \IEEEauthorblockN{Homer Simpson}
% \IEEEauthorblockA{Twentieth Century Fox\\
% Springfield, USA\\
% Email: homer@thesimpsons.com}
% \and
% \IEEEauthorblockN{James Kirk\\ and Montgomery Scott}
% \IEEEauthorblockA{Starfleet Academy\\
% San Francisco, California 96678--2391\\
% Telephone: (800) 555--1212\\
% Fax: (888) 555--1212}
% }

% conference papers do not typically use \thanks and this command
% is locked out in conference mode. If really needed, such as for
% the acknowledgment of grants, issue a \IEEEoverridecommandlockouts
% after \documentclass

% for over three affiliations, or if they all won't fit within the width
% of the page, use this alternative format:
% 
%\author{\IEEEauthorblockN{Michael Shell\IEEEauthorrefmark{1},
%Homer Simpson\IEEEauthorrefmark{2},
%James Kirk\IEEEauthorrefmark{3}, 
%Montgomery Scott\IEEEauthorrefmark{3} and
%Eldon Tyrell\IEEEauthorrefmark{4}}
%\IEEEauthorblockA{\IEEEauthorrefmark{1}School of Electrical and Computer Engineering\\
%Georgia Institute of Technology,
%Atlanta, Georgia 30332--0250\\ Email: see http://www.michaelshell.org/contact.html}
%\IEEEauthorblockA{\IEEEauthorrefmark{2}Twentieth Century Fox, Springfield, USA\\
%Email: homer@thesimpsons.com}
%\IEEEauthorblockA{\IEEEauthorrefmark{3}Starfleet Academy, San Francisco, California 96678-2391\\
%Telephone: (800) 555--1212, Fax: (888) 555--1212}
%\IEEEauthorblockA{\IEEEauthorrefmark{4}Tyrell Inc., 123 Replicant Street, Los Angeles, California 90210--4321}}

% use for special paper notices
%\IEEEspecialpapernotice{(Invited Paper)}

% make the title area
\maketitle

% As a general rule, do not put math, special symbols or citations
% in the abstract
\begin{abstract}
Many image transformations can be modeled by image operators that are
characterized by pixel-wise local functions defined on a finite
support window. In image operator learning, these functions are estimated
from training data using machine learning techniques. Input size is
usually a critical issue when using learning algorithms, and it limits
the size of practicable windows. We propose the use of convolutional
neural networks (CNNs) to overcome this limitation. The problem of
removing staff-lines in music score images is chosen to evaluate the
effects of window and convolutional mask sizes on the learned
image operator performance. Results show that the CNN based solution
outperforms previous ones obtained using conventional learning
algorithms or heuristic algorithms, indicating the potential of CNNs
as base classifiers in image operator learning. The implementations
will be made available on the TRIOSlib project site.
\end{abstract}

% no keywords

% Scale-Aware Pixelwise Object Proposal Networks
% For peer review papers, you can put extra information on the cover
% page as needed:
% \ifCLASSOPTIONpeerreview
% \begin{center} \bfseries EDICS Category: 3-BBND \end{center}
% \fi
%
% For peerreview papers, this IEEEtran command inserts a page break and
% creates the second title. It will be ignored for other modes.
\IEEEpeerreviewmaketitle

\section{Introduction}

\begin{figure}[htb]
  \centering
  
  \includegraphics[width=0.6\linewidth]{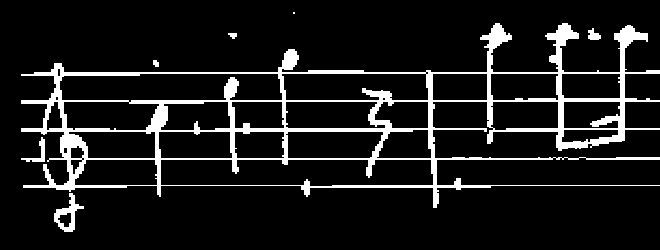}

  \medskip
  \includegraphics[width=0.6\linewidth]{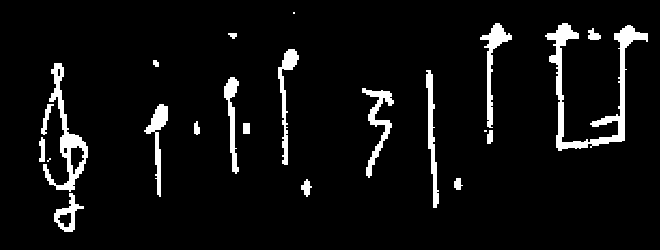}
 
\caption{\label{fig:training_image1}An example of input-output pair of
  images expressing the staff-line removal task.}
\end{figure}

\begin{figure}[htb]
  \centering

  \includegraphics[width=0.45\linewidth]{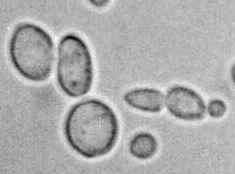}
  \ \ \  \includegraphics[width=0.45\linewidth]{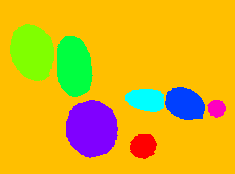}
\caption{\label{fig:training_image2}An example of input-output pair of
  images expressing the cell segmentation task.}
\end{figure}

Image analysis and image understanding include basic image processing
tasks such as segmentation and object detection. Many advances have
been achieved recently in object detection with the use of supervised
learning techniques, and particularly of convolutional neural
networks~\cite{Krizhevsky:2012,FastRCNN}. In object recognition,
supervision is usually provided as labels attached to images or to
regions in the images. In contrast, more lower level tasks such as
segmentation require labels at pixel or superpixel
levels~\cite{Pinheiro2015FromIT}. For these type of tasks, the
simplest way of providing supervision is by means of pairs of
input-output images.
For instance, Fig.~\ref{fig:training_image1} shows a sample of input
and output images for the staff line removal problem and 
Fig.~\ref{fig:training_image2} shows an example for the cell
segmentation problem. They illustrate respectively a binary
segmentation and a multi-class segmentation problems. In both cases
the pixel value in the output image can be taken as a pixel class label.
There is a subtle difference in the nature of the transformation. In
the first example, removal of staff lines can be seen as a component
filtering transformation, resulting in binary output images (both
input and output are images of the same nature). In the second case,
although the output is still an image, its pixels values are labels
and thus of a different nature of the input image. However, multiclass
segmentation can be cast as a component detection plus component
labeling problem, and its component detection part can be essentially
represented by means of binary output images.

In this work we address problems where supervision is provided at the
pixel level and the task is modeled as an image transformation. 
%% Noise filtering in images~\cite{CoyleLin:88,Yoo:99}, which was a
%% common task some decades ago, also falls within the same class of problems.
The approach proposed here is built on the image operator learning
framework used along the years to address this type of
problems~\cite{2016:tutorialSIB,1997:BarDoughTom,2009:Nina}. This
framework uses image operators that are translation-invariant and
locally defined with respect to a finite non-empty window to model
image transformations. Estimating these type of image operators from
training data has been the subject of study since the
1990s~\cite{Dough:92a,1997:BarDoughTom,Yoo:99}, with some interesting
results in binary image processing~\cite{2009:Nina,2017:IgorPR}.
Although here we restrict ourselves to binary image transformations
(like the one in the first example above), modeling and concepts apply
also to gray-scale image transformations. Application examples
include documents~\cite{2000:hirbarter,2016:tutorialSIB},
comics~\cite{2016:ninamanpu}, noise
filtering~\cite{CoyleLin:88,Yoo:99}, retinal
images~\cite{Benalcazar2013,2016:IgorSIB}, diagrams~\cite{2009:Nina}, and others.
%%  In
%% these works, morphological representation is adopted to model the
%% image operators and statistical estimation and machine learning
%% techniques are employed to estimate an operator from the training
%% data.

The problem of learning image operators is modeled as a
problem of learning local transformations. While most earlier
approaches valued morphological representation to favor
interpretation~\cite{Dough:92a,1997:BarDoughTom,2009:Nina}, more
recent approaches~\cite{Benalcazar2013,2016:IgorICIP,2016:IgorSIB} drop the concern
of explicitly keeping the morphological representation to favor
efficiency provided by recent advances in machine learning techniques.

Still, a major limitation of most machine learning algorithms is input
size, which in the context of image operator learning is determined
by the window size. CNNs, on the other hand, can be trained with very large
inputs. Therefore, in this work we propose the use of CNNs in the
image operator learning framework. We are specially interested on how
effective CNNs are when used in the context described above. As CNNs
require adjustments of several parameters, another issue of interest
is how much effort is required to tune the parameters.
To address these issues, we consider the problem of removing
staff-lines from music score images. This problem has been receiving
attention recently~\cite{CompoICDAR2013,Dalitz08,Calvo-Zaragoza2016}
and there is a large public dataset with ground-truth
information~\cite{CompoICDAR2013}, making it interesting to our study.
Moreover, there are performance reports of heuristic methods and also
of learned image operators~\cite{2017:IgorPR}, which will allow us to
make a direct comparison.

The rest of the paper is organized as follows. In
Section~\ref{sec:background} we describe the image operator learning
framework~\cite{2016:tutorialSIB}, providing a brief but comprehensive
overview. This overview will help to place the contributions of
this work properly within the framework. In Section~\ref{sec:proposal}
we discuss how to couple CNNs to the above mentioned learning
framework. In particular, we will define an experimental protocol to
be followed to adjust the many parameters of a CNN model. Then, in
Section~\ref{sec:experiments} we describe results on parameter
determination and performance of the learned image operators. In
Section~\ref{sec:conclusion} we present the conclusions of this work
and comment on future works.

\section{Background}
\label{sec:background}

Although in practice images are defined on a finite support, to
present some definitions and notations we consider they are defined on
$\mathbb{E}=\mathbb{Z}^2$. A binary image defined on $\mathbb{E}$ is a
function $f: \mathbb{E} \to \{0,1\}$, which can be represented
equivalently by the set $S_f = \{\, p \in\mathbb{E} \,:\, f(p) = 1
\}$. To simplify
notation, we will use the same symbol $S$ to denote a binary image
both as a function and as a set. Thus, $S(p)=1$ and $p \in S$ has the
same meaning. They both mean that the value of the image at point $p$
is 1. The collection of all binary images on $\mathbb{E}$ will be
denoted $\mathcal{P}(\mathbb{E})$.

A binary image operator $\Psi: \mathcal{P}(\mathbb{E}) \to
\mathcal{P}(\mathbb{E})$ that satisfies translation-invariance and
local definition w.r.t. a finite non-empty window $W$ (containing the
origin of $\mathbb{E}$) can be expressed as
\begin{equation}
  \label{eq:local_definition}
  [\Psi(S)](p) = \psi(S_{-p}\cap W)
\end{equation}
where $\psi$ is a function from $\mathcal{P}(W)$ to $\{0,1\}$ and
$S_{-p}$ is the translation of image $S$ by $-p$.

This means that operators that satisfy both properties are fully
defined by its characteristic function $\psi$. An interesting
consequence of this property is that the problem of learning an image
operator can be reduced to the problem of learning the characteristic
function $\psi$~\cite{2016:tutorialSIB}.

Given an input-output pair $(I,O)$ of training images, training data
consists of image patches $I \cap W_{p}$, collected from $I$ by
sliding window $W$ over every pixel $p$, together with its respective
label $O(p)$. In practice, image patches are flattened and their
vectorial form are used as inputs for training and 
classification~\cite{2016:tutorialSIB}.

Once a local function is learned, one can compute the output image by
applying Eq.~\ref{eq:local_definition} on every point of an input test
image. Given image pairs $(I_i,O_i)$, $i=1,\ldots,N$, the empirical
mean absolute error (MAE) of $\psi$ is defined as
\begin{equation}
  \label{eq:MAE}
  Err(\psi) = \frac{1}{T}\sum_{i=1}^{N} \sum_{p \in D(I_i)}
  |\psi([I_i]_{-p}\cap W) - O_i(p)| 
\end{equation}
where $D(I)$ denotes the support of $I$ and $T$ the total of points
considered in the summation. This error is equivalent to the
pixel-wise accuracy.

Most previous works related to image operator learning use 
approaches that preserve the morphological representation of
$\psi$. In the case of binary
images, local functions are logical functions and they can be
expressed as a sum of product
terms~\cite{1997:BarDoughTom,2009:Nina,2016:tutorialSIB}. Their
corresponding morphological
representation allows one to understand the learned operator in terms
of basic morphological operators such as erosions, dilations and
interval operators~\cite{Heij:94}.

As window points can be regarded as variables (or features), there is
a tradeoff between window size and generalization performance. Too
small windows do not suffer from generalization error but their
discriminatory power is limited. On the other hand, albeit having good
discriminatory power, too large windows
lead to poor generalization (small training error but large test
error)~\cite{2016:tutorialSIB}.

One technique to overcome this limitation to learn operators defined
w.r.t. larger windows is the two-level design
approach~\cite{2009:Nina}. This approach
consists in designing multiple operators on moderate size windows and
then taking the responses of each of them as second level features,
which are then used to train a combiner. The combiner is, ultimately,
an image operator that is locally defined w.r.t. a larger window
(the union of all moderate size windows used in the first-level
learning phase). Empirical analysis indicate that combinations
consistently provide better results than single operators.
%Both single level and two-level operators are equivalent to Boolean
%functions (logic functions).
A drawback of these approaches that are
based on explicit morphological representation is the size complexity
of such representations. The number of required elementary operators
may be exponential to the window size~\cite{2016:tutorialSIB}.

More recent approaches take advantage of modeling and processing
flexibility provided by modern machine learning
algorithms~\cite{2016:IgorICIP,2016:IgorSIB} and do not consider an
explicit morphological representation. This represents a view shift in
the modeling of image operator learning processes from approaches
based on estimating local functions to approaches based on classifier
learning. In classifier based approaches, image patches observed through
$W$ take the place of input vectors $\mathbf{x}$, and the respective
values in the output image take the place of class label $y$. Then,
conventional
classifier learning methods can be applied either directly on training
samples $(\mathbf{x}_i,y_i)$ or on transformed ones
$(\phi(\mathbf{x}_i),y_i)$ (where $\phi$ is a feature transformation
function).

\section{Proposed approach}
\label{sec:proposal}

We propose the use of CNNs as the classifier model to learn the local
functions in the image operator learning framework described
above. CNNs have some contrasting characteristics compared to
previously used algorithms. First, CNNs are known by their ability to
learn relevant features from data. Thus, they remove the need to
handcraft and select suitable feature mappings $\phi$ for each
application. In this sense, they form a generic method, but 
at the same time they possess the ability to encode problem specific
knowledge. Second, CNNs can handle relatively large input
images, managing generalization issues through supposedly powerful regularization
techniques. Therefore, in our context they could be applied to learn
local functions w.r.t. large windows (image patches) directly, rather
than in multiple training phases as in the two-level approach
proposed in~\cite{2009:Nina}. Next we describe a method to obtain a
CNN to be effectively used.

\subsection{Convolutional neural network architecture}

A common issue when dealing with convolutional neural 
networks is the determination of an adequate architecture. 
%During the last years, a variety of CNN architectures  
%and optimization techniques been proposed. Instead of doing an 
%exhaustive search of those architectures,
A standard building block of CNNs is composed of convolution, ReLU
and pooling layers, and a typical architecture of CNNs consists of
multiple layers of these building blocks followed by fully connected
layers. Some of the parameters that need to be defined before training
are:
\begin{itemize}
  \item input size: in our case, it is determined by the window size
  \item number of building blocks
  \item number of convolution masks in each block
  \item size of the convolution mask
  \item padding or no padding to compute the convolutions
  \item type of pooling
  \item number of fully connected layers
  \item cost function
  \item cost function optimization method, which may include dropout,
    and have varying learning rates and mini-batch sizes.
\end{itemize}

\subsection{Parameter evaluation}

To avoid trying all possibilities for the parameters, which is clearly
prohibitive, one can take the values used in similar works as reference.
Since here we are dealing with binary images, at pixel classification
level, there are only very few references. Thus, we opt to fix some of
the parameters based on a preliminary evaluation and then evaluate the
remaining parameters more carefully, as described below.

%% In this work, determine a general architecture through 
%% evaluation over a subset of the training set, and over that 
%% architecture, we examine the impact of varying
%% the input window size, first layer filter size, 
%% learning rate, and dropout rate.

%% The main CNN architecture consisted of two sequences of
%% convolution maps, Relu, and max-pooling; followed by two fully
%% connected layers, with dropout regularization in the penultimate
%% layer. The number of input units, or features, varied according to the 
%% window size. At this regard, we considered only square windows, 
%% varying from $9\times9$ (81 features) to $19\times19$ (361 features).
%% The number of the first layer convolution maps was fixed to $32$.
%% For optimization, we used the Adam algorithm~\cite{Kingma:2014}
%% with stochastic gradient descend and mini-batch size of $50$. 
%% More details about the architecture can be found in the code released as
%% open-source~\footnote{link omitted.}. 

We have fixed a basic architecture consisting of two building blocks
(i.e., convolution-ReLU-pooling layers) followed by two fully connected
layers, and \textit{softmax} output. The number of convolutions in
each block is fixed to 32, and the mask size to $5 \times 5$. We also
adopt zero padding, stride 1 and max-pooling of $2\times 2$. For training, we use cross-entropy
cost function along with the Adam algorithm~\cite{Kingma:2014}, and
stochastic gradient descend with mini-batch size of $50$ for
optimization. We apply dropout regularization in the penultimate
layer.
% More details about the architecture can be found in the open-source
% project~\footnote{\url{https://github.com/fjulca-aguilar/DeepTRIOS}.}.

Given the fixed configuration above, we define an experimental
protocol to evaluate the input window size, learning rate, and dropout
rate. Note that the input window determines the image patches (raw
input features) that will be used as input for training and
classification of pixels. While most applications of CNN in the
computer vision field consider a fixed input image size, window size
is an important parameter in image operator learning~\cite{2009:Nina}.

\begin{figure*}[htb]
\centering
\includegraphics[width=0.75\linewidth]{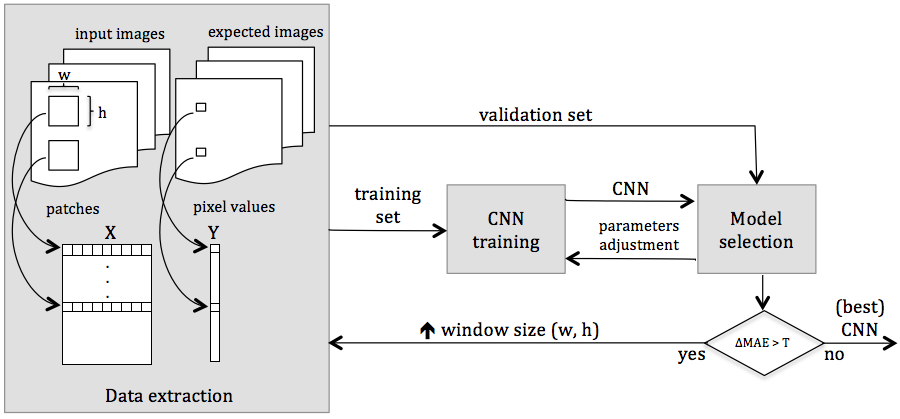}
\caption{CNN training and parameters evaluation. The input window size
  determines the patch size. We train and evaluate multiple CNN models
  over incremental window sizes. At each evaluated window size, we use
  the optimal parameters from smaller windows to narrow the parameters
  range.}
\label{fig:training} 
\end{figure*}

The strategy we follow to tune the CNN parameters is illustrated in 
Figure~\ref{fig:training}. For each window we evaluate multiple
models, varying the value for the other parameters. This process is
repeated for incremental window sizes. We start with a small window
(in our case, $9\times 9$), as they are relatively faster
to train and to provide feedback on the relevant ranges for the
parameters. We then use the promising ranges to narrow the search
range for larger windows. We use a validation set to evaluate the
effects of distinct parameter values. Specifically, we have the
following steps:

\begin{enumerate}
\item \emph{Grid search of parameters:} We use the training set to
  train multiple instances of the basic CNN model above, varying the
  learning and dropout rates.
  %rates, dropout, and first layer filter sizes.
  For the smaller windows we first consider a coarse search on a wide
  range (for example, $[10, 10^{-6}]$ for the learning rate); then we
  narrow the ranges according to the best parameters found, and do a
  finer search. For the larger windows, we start from the narrowed
  search ranges. For each training instance, we run 50 epochs, 
  recording the CNN model after each epoch.

\item \emph{Model selection:} The empirical mean absolute error
  (Eq.~\ref{eq:MAE}) of the $50$ CNN models of each training instance
  is computed on the validation set. Then, for each window size, the
  model with the lowest validation error is selected as the best model
  for that window.
\end{enumerate}

We keep training CNNs over incremental window sizes until no
considerable improvements are obtained. After the best CNN models per
window are selected, we compare them and select the overall optimal
one (the one with the lowest empirical MAE on the validation set).

A particular characteristic in image operator learning, and specially
with respect to binary images, is the fact that many repeated patches
may occur; indeed we can find cases where a same input patch has
different expected labels (that is, cases where $\mathbf{x}_i =
\mathbf{x}_j$ but $y_i \neq y_j$). Therefore, the total number of
training patches may be much larger than the number of distinct
patches, which can considerably reduce the training efficiency.
Thus, for model selection we randomly selected subsets of the
training samples. After the best model is selected, we train it again  
using the whole training set.

\section{Experimentation}
\label{sec:experiments}

In this section we show the application of the proposed method (model
selection and the performance of the selected model) on the problem of
staff-line removal in music score images.

\subsection{Experimental setup}

We used the dataset of the \emph{ICDAR 2013 music scores
  competition: staff removal}~\cite{CompoICDAR2013}, as done
in~\cite{2017:IgorPR}. The dataset consists of handwritten music score
images (with dimensions around $3400\times 2300$ pixels), modified in
order to emulate degraded documents. The dataset is divided into
training and test sets. Following~\cite{2017:IgorPR}, we selected 50
images for training and 20 for validation from the provided training
set. From each image, we extracted patches centered at each white
pixel of the image\footnote{We consider only white
  pixels since the expected result is a subset of the input image. In
  terms of image operators, it is expected that the learned operator
  should be anti-extensive, i.e., that $\Psi(S) \subseteq S$. Thus,
  there is no need to estimate function value for the patches centered
  on background pixels.}.
The total number of patches extracted from the
training and validation images was about $27$ and $10$ millions,
respectively. To select the best model for each window, CNNs were
trained using $5$ million patches randomly selected from the training
set and evaluated also on $5$ million patches randomly selected from
the validation set.

To evaluate the performance of the selected model, we measured
accuracy, specificity, and recall on the test set, considering only
the white pixels of the input images. In this setup, specificity
measures the effectiveness of a method to preserve non-staff line
pixels, and recall measures the effectiveness of the method to detect
staff line pixels.

For implementing the CNNs, we used the TensorFlow library
(\url{http://tensorflow.org/}) %~\cite{tensorflow:2015},
integrated with TRIOSlib, a library for image operator
learning~\cite{2016:tutorialSIB}. The implementations will be made
available on the TRIOSlib project site
(\url{http://github.com/trioslib/trios}).

\subsection{Results on model selection}

For all evaluated window sizes, best models were found using a
learning rate in the $[10^{-5}, 10^{-6}]$ range, without significant
variation when using larger or smaller windows. Also, the use of
dropout did not considerably improved performance. This might be due
to the fact that we did not use a very deep CNN.

\begin{figure}[htb]
\centering
\includegraphics[width=\linewidth]{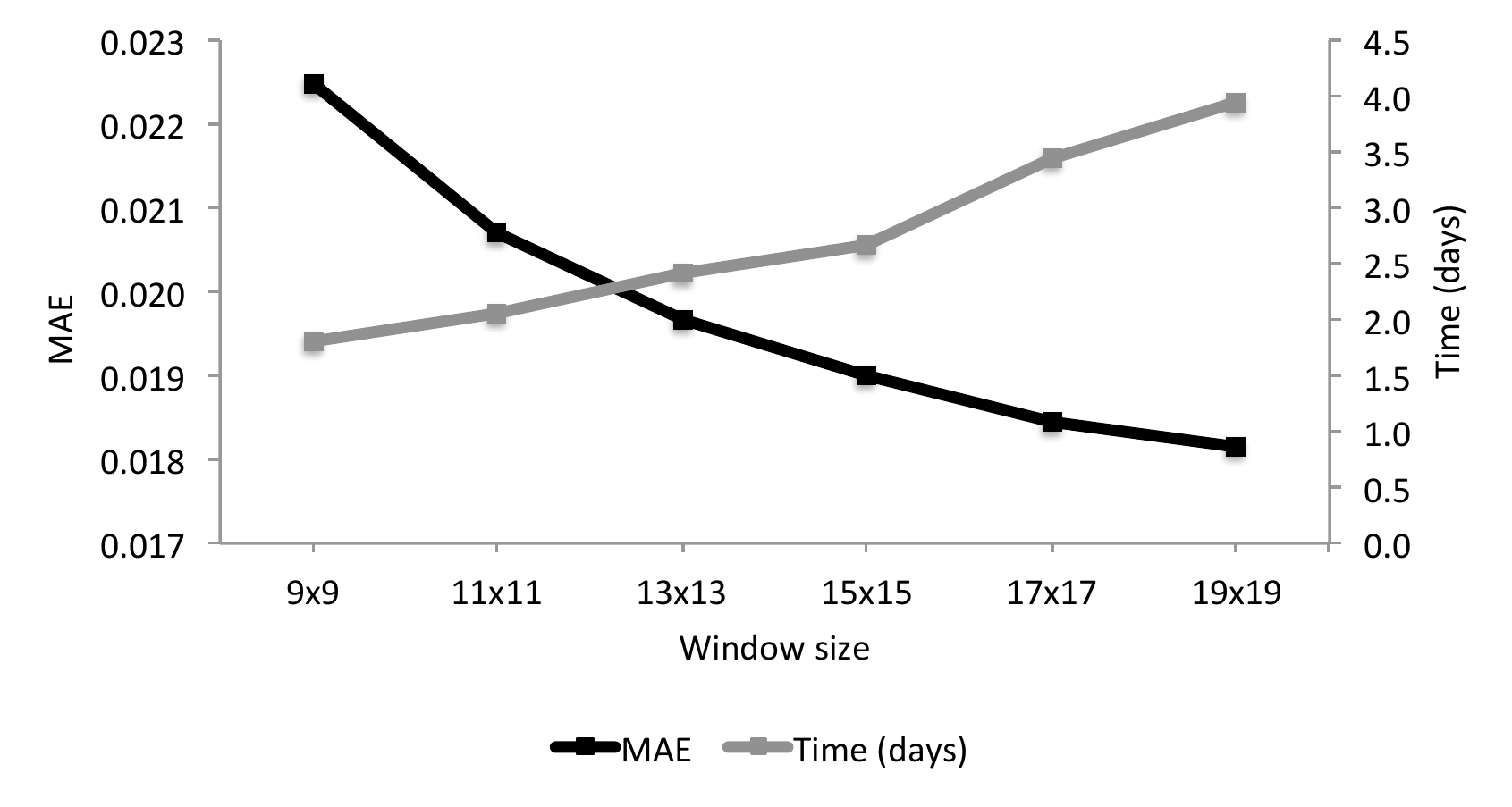}
\caption{Mean absolute error (MAE, left hand axis) and 
training time (right hand axis) for the best CNN models 
per window size.}
\label{fig:best_window} 
\end{figure}

Figure~\ref{fig:best_window} shows the MAE on validation set and
the training time of the best CNN models per window size. We can see a
small but consistent accuracy improvement, or MAE reduction, as the
window size is incremented. However, such improvement also requires an
increasing training time due to the larger input size. 
% From an application view, 
% the use of larger windows has the downside of requiring 
% considerable more training time. 
For instance, the training of all instances with the $9\times 9$
window took less than one day while with  the $19\times 19$ window
took about four days. All the experimentation was done on CPU only.

\begin{figure*}[!t]
\centering
\subfloat[]{\includegraphics[width=0.3\linewidth]{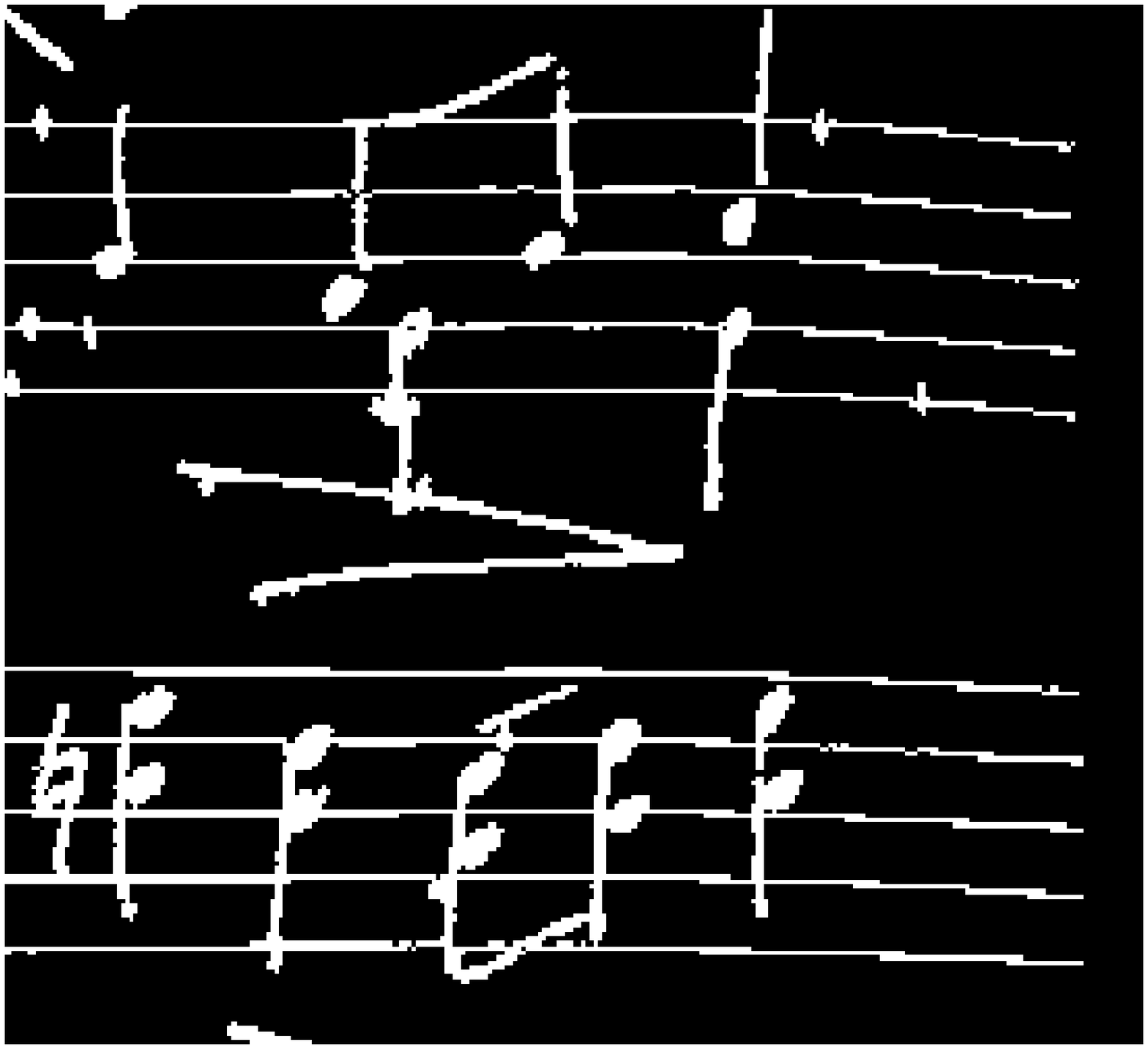}%
\label{fig:BW_1593_segment}}
\hspace{3em}
\subfloat[]{\includegraphics[width=0.3\linewidth]{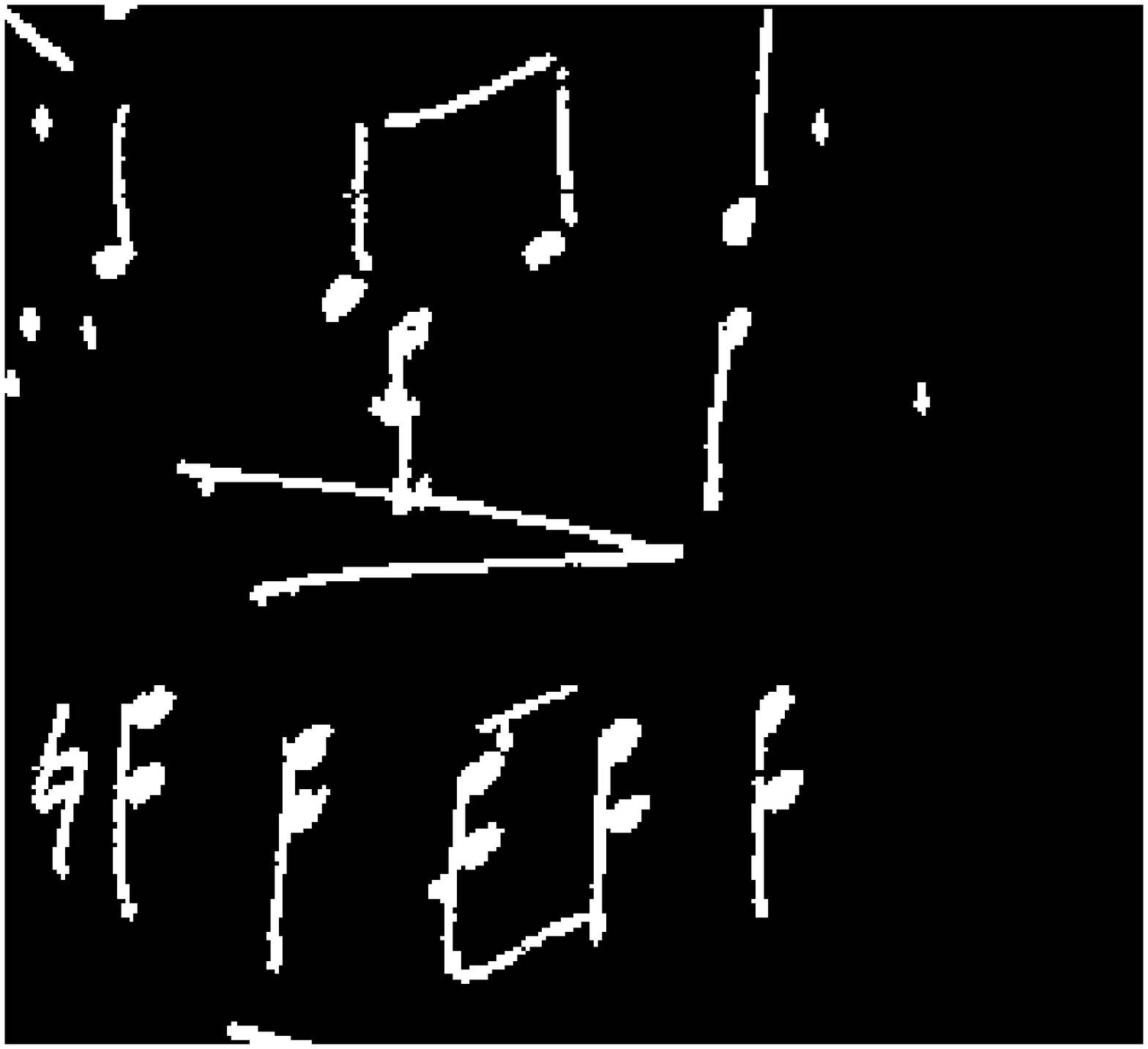}%
\label{fig:GT_1593_segment}}
% \hfill
% \hspace{3em}
\\

\subfloat[]{\includegraphics[width=0.3\linewidth]{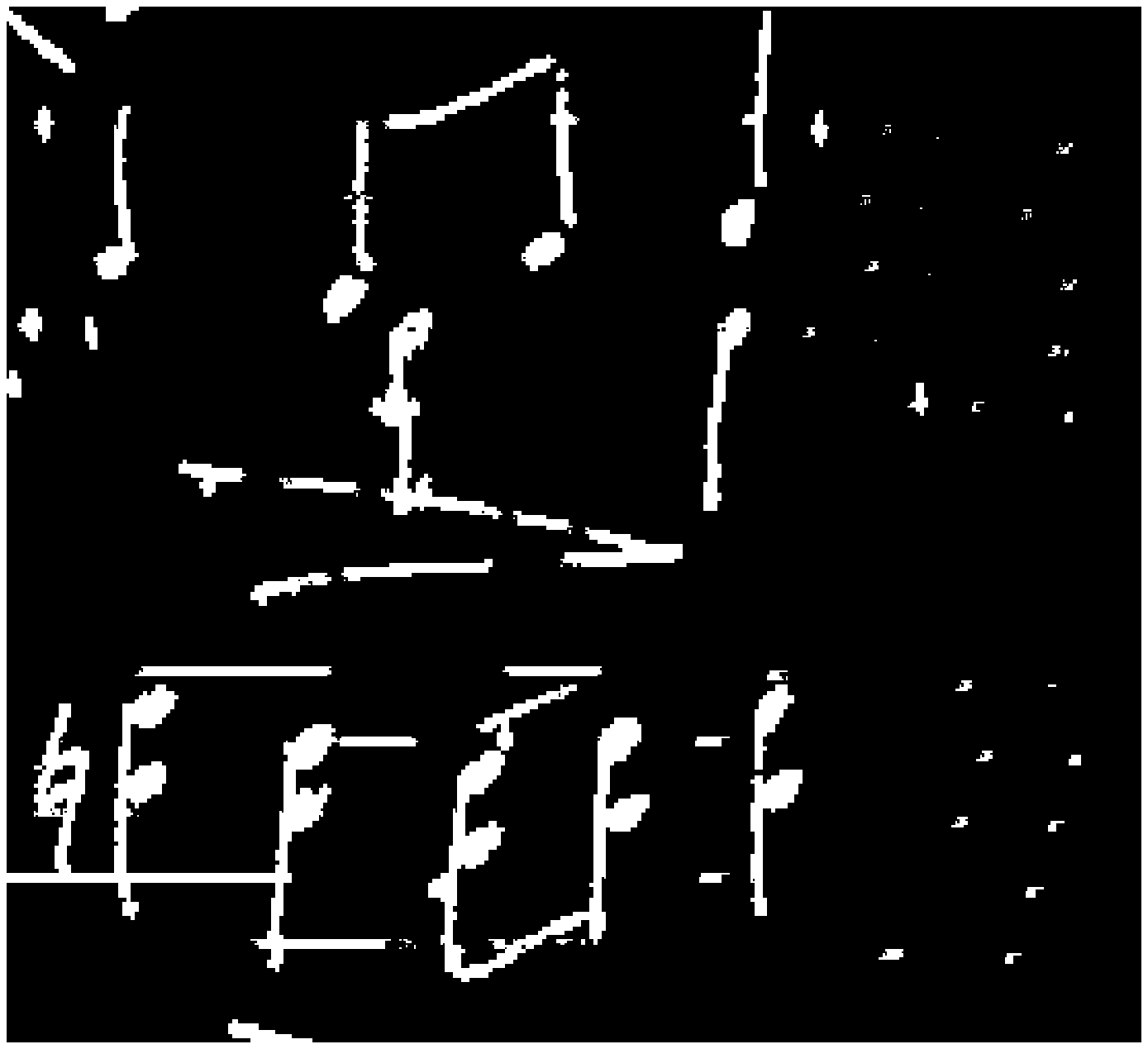}%
\label{fig:9x9_segment}}
% \hfill
\hspace{3em}
\subfloat[]{\includegraphics[width=0.3\linewidth]{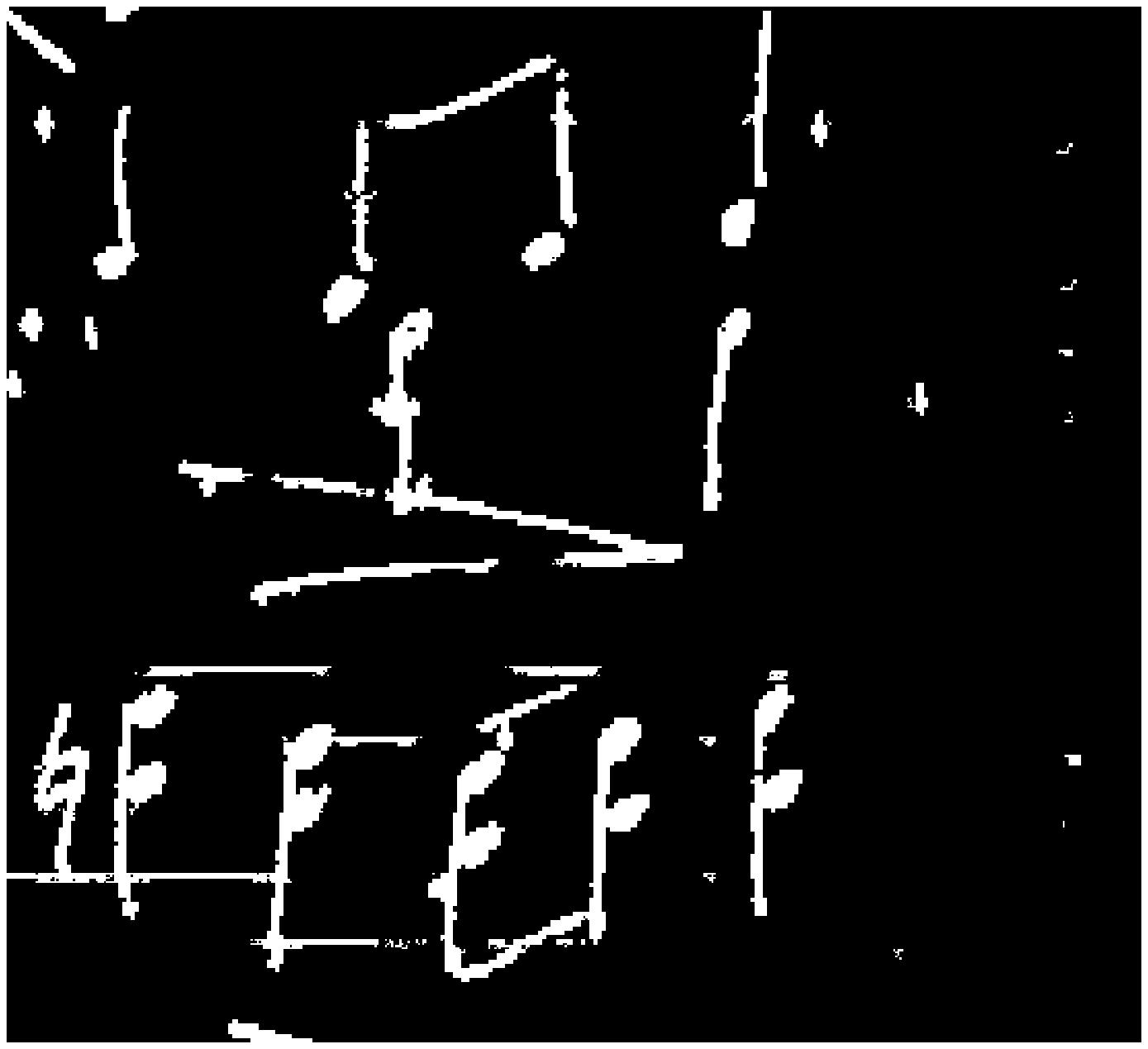}%
\label{fig:19x19_segment}}
\caption{Resulting image analysis: (a) region of an image for which
  CNNs obtained worst accuracy, (b) same region in the expected output
  image, and same region in the outputs produced by the best CNN
  models with (c) $9\times9$ and (d) $19\times 19$ windows.}
\label{fig:worst_images}
\end{figure*}

To understand the errors, we analyzed images with the largest
errors. Figure~\ref{fig:worst_images} shows a region of one of such
images, the same region in the expected output, and in the results
obtained using the best CNN models with the $9\times9$ and the
$19\times19$ windows. We have noticed that a common error is the
misclassification of pixels of thick staff line segments as non-staff
line pixels (the horizontal lines at the bottom-left part of the
figures). Another type of error is on the extremities of the
staff-lines.
% In the latter, the thick segments are generated due to small changes
% of direction of the lines, which in handwriting often occurs at the
% end of strokes (hooks).
It can be seen that the errors are more evident in the output of the
CNN with the smaller window.

\begin{figure}[htb]
\centering
\includegraphics[width=\linewidth]{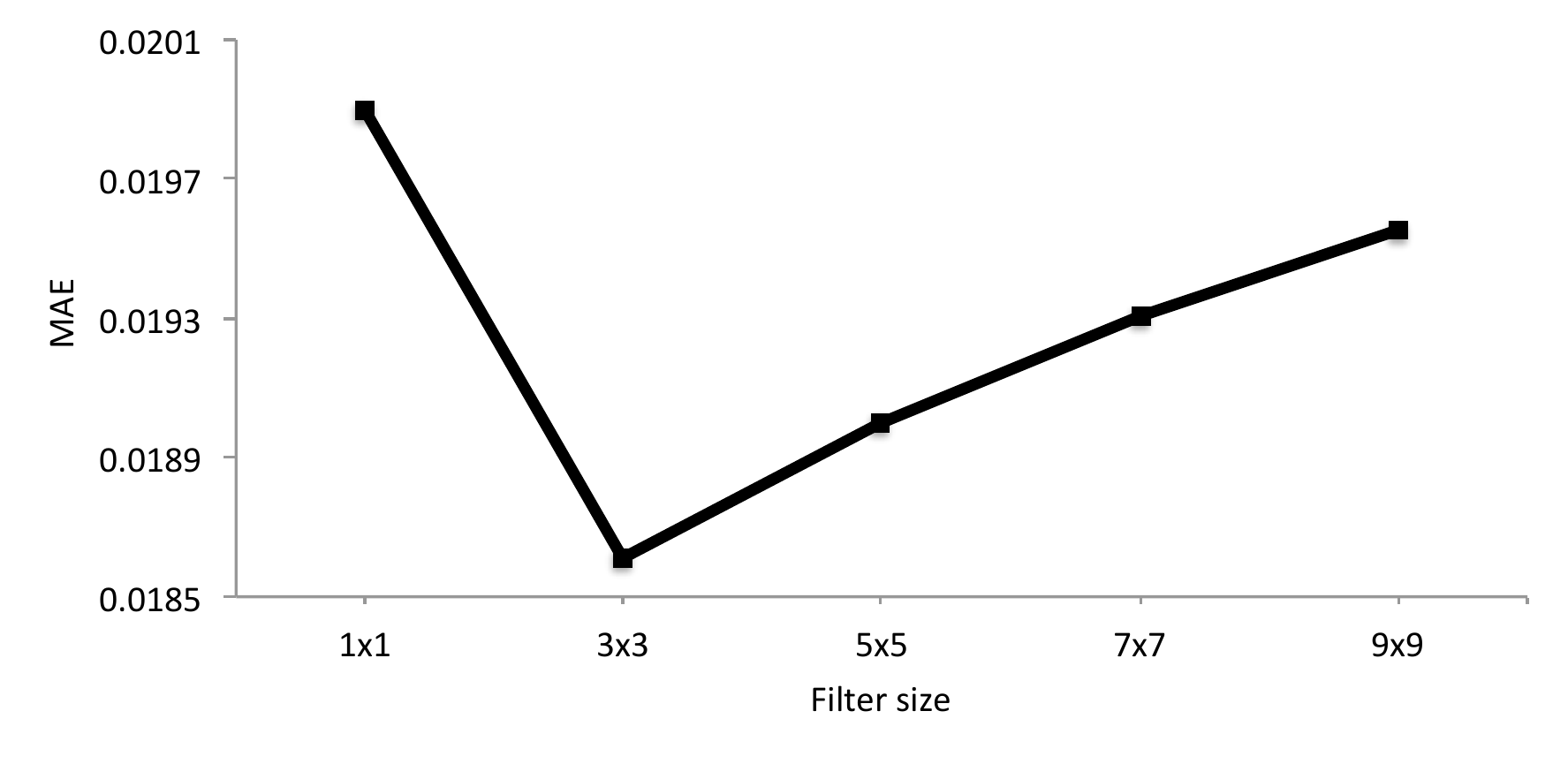}
\caption{Mean absolute error (MAE) on validation set for 
different first layer filter sizes (window size fixed to $15 \times 15$).}
\label{fig:best_filter_size}
\end{figure}

The convolution mask size was fixed to $5\times 5$ during model
selection. As we intuitively reasoned that larger filters would allow
CNNs to better capture staff line patterns, we did an experiment to
evaluate this particular parameter for the case of $15\times 15$
window. We have found that higher accuracies are achieved with the
$3\times 3$ mask, as shown in Figure~\ref{fig:best_filter_size}.
This indicates that additional study is necessary to undertand the
effects of the convolution mask sizes on the results. 

\subsection{Performance of the selected model}

According to the experiments in the model selection step, the best
overall model was the best one in the group of those trained with
$19\times 19$ window. The best model was retrained on the whole
training set and evaluated on the test set.  
Table~\ref{tab:test_results} shows the accuracy, specificity, and
recall of the selected model as well as the ones provided
in~\cite{2017:IgorPR}, which include results of four state-of-the-art
methods. Three of the methods (LTC, Skeleton, LRDE) consist of
heuristic techniques and the other (FS-MI) consists of a two-level
operator learning approach with automatic window determination.
Our method outperforms previous methods in terms of accuracy and
recall, and also obtains the third best result in terms of
specificity.

\begin{table}[h]
\caption{Performance comparison ($\%$) of our method with previous
  methods (performance of other methods extracted from~\cite{2017:IgorPR}).}
\begin{center}
\begin{tabular}{lccc}
\hline
Method & Accuracy & Specificity & Recall\\ \hline 
LTC &  87.58 & \textbf{99.52} & 67.76\\
Skeleton &  94.50 & 99.03 & 86.97\\
LRDE &  97.03 & 98.84 & 94.02 \\
FS-MI &  96.96 & 98.46 & 94.48\\
Our method &  \textbf{97.96} & 98.98 & \textbf{95.72} \\ \hline
\end{tabular}
\end{center}
\label{tab:test_results}
\end{table}

\subsection{Discussion}

\begin{figure}[htb]
\centering
\includegraphics[width=0.45\linewidth]{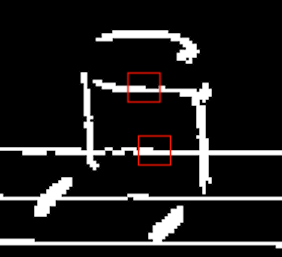}
\caption{Some pixels might be difficult to classify by looking 
only at a local patch. The two highlighted patches are of size
$19\times 19$ and are very similar each other, but they are in
distinct classes.}
\label{fig:horizontal}
\end{figure}

Although CNNs with larger windows have shown improved accuracy 
in comparison to CNNs with smaller windows, some pixels 
might not be correctly classified even using large windows.
For instance, Figure~\ref{fig:horizontal} highlights two patches. The
top one is on a beam note, capturing a horizontal line segment of the
symbol. The second is centered on a staff line pixel. Visually they
are very similar and the window required to distinguish them may be
too large.
%% An example is a pixel at thick staff lines, 
%% which might be interpreted by the classifier as 
%% pixels of breath marks, or horizontal segments of 
%% musical symbols. 
A post processing of the output image might be an option to avoid
using too large windows.

Our best CNN model uses 361 features ($19\times 19$ windows),
representing a large increment in the number of features compared to
the classifiers used in~\cite{2017:IgorPR} (40 features within a
$13\times 17$ window domain). One advantage of the image learning
framework is the fact that it is data-driven, adapting to the
particularities of the considered family of images. This flexibility
plus the ability of CNN on working with large inputs might be the
explanation of superior results compared to previous methods. 
%% These two points, along with the availability
%% of a large training data set, explain, at some extent, the better CNN
%% generalization.

\section{Conclusion}
\label{sec:conclusion}

We introduced CNNs as local classifiers in the image operator learning
framework. CNNs overcome limitations of previous methods with regard
to practicable window sizes. In addition, CNNs remove the need of
manually selecting features or combining simpler operators. This
generality and flexibility comes accompanied with the challenge of
finding an adequate set of CNN hyperparameters. We showed that by
using standard architecture and parameter optimization methods, we
obtain a CNN-based image operator that outperforms state-of-the-art
staff line removal methods. Hence, we conclude that the use of CNNs as
base classifiers in the image operator learning framework opens a
promising path.
%, based on the encouraging results and the expectation that further
%tuning of the parameters may lead to better performance.

We note, however, that while the above statement fits well in
scenarios where there is abundance of training data, there is still
few knowledge regarding scenarios with few training data. Some of the
issues to be further investigated include the application of the
proposed method on images of different domains and also on gray-scale
images. It would be interesting to test the limits of CNN (for
instance, how far accuracy improvement can be pushed by augmenting the
number of convolutional layers or window size?), adapt previously
trained CNNs to other types of transformations, and understand which
are the features extracted by CNNs.

% use section* for acknowledgment
\section*{Acknowledgment}
F.~D.~Julca-Aguilar thanks FAPESP (grant
2016/06020-1). N.~S.~T.~Hirata thanks CNPq (305055/2015-1).
This work is supported by FAPESP (grant 2015/17741-9) and CNPq (grant
484572/2013-0).

% trigger a \newpage just before the given reference
% number - used to balance the columns on the last page
% adjust value as needed - may need to be readjusted if
% the document is modified later
%\IEEEtriggeratref{8}
% The "triggered" command can be changed if desired:
%\IEEEtriggercmd{\enlargethispage{-5in}}

% references section

% can use a bibliography generated by BibTeX as a .bbl file
% BibTeX documentation can be easily obtained at:
% http://mirror.ctan.org/biblio/bibtex/contrib/doc/
% The IEEEtran BibTeX style support page is at:
% http://www.michaelshell.org/tex/ieeetran/bibtex/
\bibliographystyle{IEEEtran}
\bibliography{refs}

% that's all folks
\end{document}